\documentclass[orivec,runningheads]{llncs}
\usepackage{times}

\usepackage[defblank]{paralist}
\usepackage{enumerate}

\usepackage[hidelinks]{hyperref}
\usepackage[english]{cleveref}
\crefformat{footnote}{#2\footnotemark[#1]#3}
\usepackage[utf8]{inputenc}
\usepackage{amsmath,amssymb}
\usepackage{xspace}
\usepackage{graphicx}

\usepackage{algpseudocode}
\usepackage{algorithm}
\usepackage{algorithmicx}

\hypersetup{%
    pdftitle={Specification-Driven Multi-Perspective Predictive
      Business Process Monitoring (Extended Version)},%
    pdfauthor={Ario Santoso, ORCID of the author: http://orcid.org/0000-0001-6212-4365},%
    pdfsubject={Specification-Driven Multi-Perspective Predictive
      Business Process Monitoring (Extended Version)},%
    pdfkeywords={Predictive Business Process Monitoring, Prediction
      Task Specification, Automatic Prediction Model Creation, Multi-perspective Prediction},%
    pdfcreator={Ario Santoso, ORCID of the author: http://orcid.org/0000-0001-6212-4365},%
}

\graphicspath{{figures/}}

\newcommand{\set}[1]{\{#1\}}                      

\newcommand{\card}[1]{|{#1}|}                     

\newcommand{\tup}[1]{\langle#1\rangle}            

\newcommand{\A}{\mathcal{\uppercase{A}}}

\newcommand{\E}{\mathcal{\uppercase{E}}}

\renewcommand{\P}{\mathcal{\uppercase{P}}}

\newcommand{\R}{\mathcal{\uppercase{R}}}

\newcommand{\ra}{\rightarrow}

\newcommand{\abssep}{\ensuremath{\mathbf{\cdot~}}}

\newcommand{\udefined}{\bot}

\newcommand{\seq}{\sigma}

\newcommand{\prefix}[2]{#2^{#1}}

\newcommand{\eventuniv}{\ensuremath{\E}\xspace}
\newcommand{\attnameuniv}{\ensuremath{\A}\xspace}

\newcommand{\attval}[1]{\#_{\text{{#1}}}}

\newcommand{\event}{\ensuremath{e}\xspace}
\newcommand{\trace}{\ensuremath{\tau}\xspace}
\newcommand{\eventlog}{\ensuremath{L}\xspace}

\newcommand{\str}{\mathsf{String}}
\newcommand{\num}{\mathsf{number}}
\newcommand{\lcop}{\text{\textbf{lcop}}}
\newcommand{\acop}{\text{\textbf{acop}}}

\newcommand{\langname}{\text{First-Order Event Expression}\xspace}
\newcommand{\langnameabr}{\text{FOE}\xspace}

\newcommand{\true}{\mathsf{true}}
\newcommand{\false}{\mathsf{false}}

\newcommand{\fforall}{\forall}
\newcommand{\fexists}{\exists}
\newcommand{\fand}{~\wedge~}

\newcommand{\fimpl}{~\ra~}

\newcommand{\fandcompact}{\wedge}
\newcommand{\forcompact}{\vee}
\newcommand{\fimplcompact}{\ra}

\newcommand{\fsum}{{ \texttt{SUM}}}
\newcommand{\fconcat}{{ \texttt{CONCAT}}}

\newcommand{\eventquery}[2]{\text{\texttt{\small e}}[#1]\textbf{. }{\text{{#2}}}}

\newcommand{\eventexpshort}{\ensuremath{\mathsf{eventExp}}\xspace}
\newcommand{\numexpb}{\ensuremath{\mathsf{numExp}}\xspace}
\newcommand{\nonnumexpb}{\ensuremath{\mathsf{nonNumExp}}\xspace}

\newcommand{\noteq}{\neq}
\newcommand{\eq}{=}

\newcommand{\idx}{\mathsf{idx}\xspace}
\newcommand{\nat}{\text{\textit{pint}}\xspace}

\newcommand{\last}{\ensuremath{\mathsf{last}\xspace}}
\newcommand{\curr}{\ensuremath{\mathsf{curr}\xspace}}

\newcommand{\ar}{R\xspace}
\newcommand{\arset}{\R\xspace}
\newcommand{\targetarrow}{\Longrightarrow}

\newcommand{\cond}{\mathsf{Cond}}
\newcommand{\target}{\mathsf{Target}}
\newcommand{\otarget}{\mathsf{DefaultTarget}}

\newcommand{\ruletup}[1]{\tup{#1}}

\newcommand{\analrule}{analytic rule\xspace}
\newcommand{\analrules}{analytic rules\xspace}
\newcommand{\AnalRule}{Analytic Rule\xspace}

\newcommand{\eval}{evaluate}

\newcommand{\condtargetrules}{\text{conditional-target rules}\xspace}

\newcommand{\inter}[4]{(#1)^{#2^{#3}}_{#4}}
\newcommand{\val}{\ensuremath{\nu}}

\newcommand{\sat}[3]{#1^{#2} \models #3}

\newcommand{\satisfyb}[4]{(#4)^{#1^{#2}}_{#3}}

\algnewcommand\algorithmicforeach{\textbf{for each}}
\algdef{S}[FOR]{ForEach}[1]{\algorithmicforeach\ #1\ \algorithmicdo}

\newcommand{\encfunc}{\mathsf{enc}}
\newcommand{\encset}{\mathsf{Enc}}
\newcommand{\predfunc}{\P}

\newcommand{\condpingponga}{\cond_{\text{pp}}}
\newcommand{\condpingpongb}{\cond_{\text{pp2}}}
\newcommand{\targetremtime}{\target_{\text{remainingTime}}}

\newcommand{\condpingpongteam}{\cond_{\text{ppteam}}}

\begin{document}

\mainmatter

%
%

\title{Specification-Driven Multi-Perspective \\Predictive Business
  Process Monitoring \\(Extended Version)}
\titlerunning{Specification-Driven Predictive Business
  Process Monitoring (Extended Version)}

\author{Ario Santoso \inst{1,2}} 

\authorrunning{Ario Santoso}

\institute{
Department of Computer Science, University of Innsbruck, Austria  \and
Faculty of Computer Science, Free University of Bozen-Bolzano, Italy 
  \email{ario.santoso@uibk.ac.at} }

\maketitle

\begin{abstract}

  Predictive analysis in business process monitoring aims at
  forecasting the future information of a running business
  process. The prediction is typically made based on the model
  extracted from historical process execution logs (event logs). In
  practice, different business domains might require different kinds
  of predictions. Hence, it is important to have a means for properly
  specifying the desired prediction tasks, and a mechanism to deal
  with these various 
  prediction tasks. Although there have
  been many studies in this area, they mostly focus on a specific
  prediction task. 
%
%
%
  This work introduces a language for specifying the desired
  prediction tasks, and this language allows us to express various
  kinds of prediction tasks.
%
  This work also presents a mechanism for automatically creating the
  corresponding prediction model based on the given specification.
  Thus, different from previous studies,
  our approach enables us to deal with various kinds of prediction
  tasks based on the given specification.
%
%
  A prototype implementing our approach has been developed and
  experiments using a real-life event log have been conducted.

\smallskip 
\textbf{Keywords:} Predictive Business Process
  Monitoring \abssep Prediction Task Specification \abssep 
  Automatic Prediction Model Creation \abssep Multi-perspective
  Prediction

\end{abstract}

\section{Introduction}
\label{sec:introduction}

Process mining~\cite{Aalst:2016} provides a collection of techniques
for extracting 
process-related information
%
from the logs of business process executions (event logs). One
important area in this field is
%
%
predictive business process monitoring,  
%
%
which aims at forecasting the future information of a running process
based on the models extracted from event logs.
Through predictive analysis, potential future problems can be detected
and preventive actions can be taken in order to avoid unexpected
situation (e.g., processing delay, SLA violations).
Many techniques have been proposed for tackling various 
prediction tasks such as predicting the outcomes of a
process~\cite{MFDG14,DDFT16,VDLMD15,PVWFT16}, predicting the remaining
processing time~\cite{ASS11,TVLD17,RW13,PSBD14,PSBD16}, predicting the
future events~\cite{DGMPY17,TVLD17,ERF17}, etc
(cf.~\cite{MLISFCDP15,MFE12,ERF17,SWGM14,PVFTW12,BMDB16,CDLVT15}).


In practice, different business areas might need different kinds of
prediction tasks. For instance, 
an online retail company might be interested in predicting the
processing time until an order can be delivered to the customer,
%
%
while for an insurance company, predicting
the outcomes of an insurance claim process would be interesting. On
the other hand, both of them might be interested in predicting whether
their processes comply with some business constraints (e.g., the
processing time must be finished within a certain amount of time).

When it comes to predicting the outcomes of a process or predicting an
unexpected behaviour, it is important 
to specify the desired outcomes or the unexpected
behaviour 
precisely.
For instance, in the area of customer problem management, 
to increase customer satisfaction as well as to promote efficiency, we
might be interested in predicting the possibility of ``\emph{ping-pong
behaviour}'' among the Customer Service (CS) officers while handling
the customer problems.
%
%
However, the definition of a ping-pong behaviour could be varied.
For instance, when a CS officer transfers a customer problem into
another CS officer who belongs to the same group, it can already be
considered as a ping-pong behaviour since both of them should be able
to handle the same problem.
Another possible definition would be 
when a CS officer transfers a problem into another CS
officer who has the same expertise, and the problem is transfered
back into the original CS officer.


To have 
a suitable prediction 
service for our domain,
we need 
to understand and specify the
desired prediction tasks properly. 
Thus, we need a means to express the specification.
Once we have characterized the prediction objectives and are able to
express them properly, we need a mechanism to create the
corresponding prediction model.
%
%
To automate the prediction model creation, the specification should be
machine processable.
%
As illustrated above, such specification mechanism should
also 
allow us to specify some constraints over the data, and compare some
data values at different time points. For example, to characterize the
ping-pong behaviour, one possibility is to specify the behaviour as
follows:
%
%
``\textit{there is an event at a certain time point in which the CS officer
is different with the CS officer 
in the event at the next time point, but both of them belong to the
same group}''. Note that here we need to compare the information about
the CS officer names and groups at different time points.











In this work, we tackle those problems by providing the following
contributions:
\begin{inparaenum}[\itshape (i)]
\item We introduce a 
  rich language for expressing the desired prediction tasks.
  This language allows us to specify various kinds of prediction
  tasks.
%
%
  In some sense, this language also allows us to specify how to create
  the desired prediction models based on the event logs.
%
\item We devise a mechanism for building the corresponding prediction
  model based on the given 
  specification. Once created, the prediction model can be used to
  provide predictive analysis service in business process monitoring.
\item We exhibit how our approach can be used for tackling various
  kinds of prediction tasks
  (cf.~\Cref{sec:showcase}). 
%
%
%
\item We develop a prototype that implements our approach and
  enables the automatic creation of prediction models based on the
  specified prediction objective.
\item To demonstrate the applicability of our approach, we carry out
  experiments using a real-life event log that was provided for the
  BPI Challenge 2013~\cite{BPI-13-data}.
\end{inparaenum}

 Roughly speaking, 
in our approach, we specify various desired prediction tasks by
specifying
how we want to map each (partial) business processes execution
information into the expected predicted information.
Based on this specification, we automatically train either
classification or regression models that will serve as the prediction
models.
%
%
By specifying a set of desired prediction
tasks, 
we can obtain multi-perspective prediction services that enable us to
focus on various aspects and predict various
information. 
%
%
%
Our approach is independent with respect to the
classification/regression model that is used. 
In our implementation, to get the expected quality of predictions, the
users are allowed to choose the desired classification/regression
model as well as the feature encoding mechanisms (to allow some sort
of feature engineering).
%
%
%
This paper is the extended version of \cite{AS-BPMDS-18} and it
provides supplementary materials for \cite{AS-BPMDS-18} by providing
more explanations, examples and experiments.

\section{Preliminaries}
\label{sec:preliminaries}

This section provides some background concepts for the rest of the paper.
\medskip
\noindent
\textbf{Trace, Event and Event Log. \xspace}
%
%
We follow the usual notion of event logs as in process
mining~\cite{Aalst:2016}. An event log captures historical information
about the execution of business processes.
%
In an event log, each execution of a process is represented as a
trace.
Each trace has several events, and each event in the trace captures the information
about a particular event that happens during the process execution.
%
%
Events are characterized by various attributes, e.g., \emph{timestamp}
(the time at which the event occurred).

Let $\eventuniv$ be the \emph{event universe} (i.e., the set of all
event identifiers), and $\attnameuniv$ be the set of \emph{attribute
  names}. For any event $\event \in \eventuniv$, and attribute name
$n \in \attnameuniv$, $\attval{n}(\event)$ denotes the \emph{value of
  the attribute} $n$ of $\event$. E.g., $\attval{timestamp}(\event)$
denotes the timestamp of the event $\event$. If an event $\event$ does
not have an attribute named $n$, then $\attval{n}(\event) = \udefined$
(undefined value).
A \emph{finite sequence over $\eventuniv$ of length $n$} is a mapping
$\seq: \set{1, \ldots, n} \ra \eventuniv$, and such a sequence is
represented as a tuple of elements of $\eventuniv$, i.e.,
$\seq = \tup{\event_1, \event_2, \ldots, \event_n}$
where $\event_i = \seq(i)$ for $i \in \set{1, \ldots, n}$.
The set of all \emph{finite sequences} over $\eventuniv$ is denoted by
$\eventuniv^*$.
%
%
The \emph{length} of a sequence $\seq$ is denoted by $\card{\seq}$.

A \emph{trace} $\trace$ is a finite sequence over $\eventuniv$ such
that each event $\event \in \eventuniv$ occurs at most once in
$\trace$, i.e., $\trace \in \eventuniv^{*}$ and for
$ 1 \leq i < j \leq \card{\trace}$, we have
$\trace(i) \neq \trace(j)$, where  
$\trace(i)$ refers to the \emph{event of the trace $\trace$ at the
  index $i$}.
%
%
%
Let $\trace = \tup{e_1, e_2, \ldots, e_n}$ be a trace,
$\prefix{k}{\trace} = \tup{e_1, e_2, \ldots, e_{k}}$ denotes the
\emph{$k$-length prefix} of $\trace$ (for $ 0 < k < n$).  
For example, let
$\set{e_1, e_2, e_3, e_4, e_5, e_6, e_7} \subset \eventuniv$,
$\trace = \tup{e_3, e_7, e_6, e_4, e_5} \in \eventuniv^{*}$ is an
example of a trace, $\trace(3) = e_6$, and $\prefix{2}{\trace} = \tup{e_3, e_7}$.
%
%
%
%
%
%
Finally, an \emph{event log} $\eventlog$ is a set of traces such that
each event occurs at most once in the entire log, i.e., for each
$\trace_1, \trace_2 \in \eventlog$ such that $\trace_1 \neq \trace_2$,
we have that $\trace_1 \cap \trace_2 = \emptyset$, where
$\trace_1 \cap \trace_2 = \set{\event \in \eventuniv~\mid~\exists i, j \in
  \mathbb{Z}^+ \text{ . } \trace_1(i) = \trace_2(j) = e }$.



%
An IEEE standard for representing event logs, called XES (eXtensible
Event Stream), has been introduced in~\cite{IEEE-XES:2016}. The
standard defines the XML format for organizing the structure of
traces, events and attributes in event logs. It also introduces some
extensions that define some attributes with pre-defined meaning such
as:
\begin{inparaenum}[\itshape (i)]
\item ``\textit{concept:name}'', which stores the name of event/trace;
\item ``\textit{org:resource}'', which stores the name/identifier of the
  resource that triggered the event (e.g., a person name);
\item ``\textit{org:group}'', which stores the group name of the resource that
  triggered the event.
\end{inparaenum}


\medskip
\noindent
\textbf{Classification and Regression. \xspace}
%
In machine learning \cite{MRT12}, a classification and regression
model can be seen as a function $f: \vec{X} \ra Y$ that takes some
\emph{input features}/\emph{variables} $\vec{x} \in \vec{X}$ and
predicts the corresponding \emph{target value/output} $y \in Y$.
%
%
The key difference is that the output range of the classification task
is a finite number of discrete categories (qualitative outputs) while
the output range of the regression task is continous values
(quantitative outputs) \cite{HPK11,FHT01}. Both of them are supervised
machine learning techniques where the models are trained with labelled
data. I.e., the inputs for the training are the pairs of input
variables $\vec{x}$ and target value $y$. This way, the models learn
how to map certain inputs $\vec{x}$ into the expected target value
$y$.


\section{Approach}
\label{sec:approach}

Our approach for obtaining a predictive process
monitoring service consists of the following main steps:
\begin{inparaenum}[\itshape (i)]
\item specify the desired prediction tasks and
\item automatically create the prediction model based on the given
  specification.
\end{inparaenum}
Once created, we can use the models to predict the future
information. In the following, we elaborate these steps.


\subsection{Specifying the Desired Prediction Tasks }\label{subsec:spec-language}

This section explains the mechanism for specifying the desired
prediction task. 
Here we introduce a language that is able to capture the desired
prediction task in terms of the specification on 
how to map each (partial) trace in the event log into the desired
prediction results.
%
%
Such specification can be used 
to train a classification/regression model that will be used as the
prediction model.


In our approach, the specification of a particular prediction task is
specified as an \analrule, where an \emph{\analrule} $\ar$ is an
expression of the form
\begin{center}
$
  \ar = \ruletup{ 
    \cond_1 \targetarrow \target_1, ~
    \ldots, ~
    \cond_n \targetarrow \target_n, ~
    \otarget
  }.
$
\end{center}
%
%
%
Each $\cond_i$ in $\ar$ is called \emph{condition expression}, while
$\target_i$ and $\otarget$ are called \emph{target expression} (for
$i \in \set{1,\ldots,n}$). 
%
%
We explain and formalize how to specify a condition and target
expression after providing some intuitions below.
%

An \analrule $\ar$ will be interpreted as a function
that maps (partial) traces into the values obtained from evaluating
the target expressions. The mapping is based on the condition that is
satisfied by the corresponding trace. 
Let $\trace$ be a (partial) trace, such function $\ar$ can be
illustrated as follows (
the formal definition will be given
later): 
%
%
%
%
\begin{center}
$
\ar(\trace)
  = \left\{ \begin{array}{l@{ \qquad}l}
\eval(\target_1) &\mbox{ if  } \trace \mbox{  satisfies } \cond_1 \mbox{, }\\
%
%
\ \ \ \ \ \ \ \ \ \vdots& \ \ \ \ \ \ \ \  \ \ \ \ \ \ \ \ \vdots\\
\eval(\target_n) &\mbox{ if  } \trace \mbox{  satisfies } \cond_n \mbox{, }\\
\eval(\otarget) & \mbox{ otherwise}
          \end{array}
\right.
$
\end{center}
We will see that a target expression essentially 
specifies the desired prediction result or expresses the way how to
compute the desired prediction result. Thus, an \analrule $\ar$ can
also be seen as a means to map (partial) traces into the desired
prediction results, or to compute the expected prediction results of
(partial) traces.


%
To specify a condition expression in \analrules, we introduce a
language called \langname (\langnameabr). Roughly speaking, an
\langnameabr formula is a First-Order Logic (FOL) formula
\cite{Smul68} where the atoms are expressions over some event
attribute values and some comparison operators (e.g., $\eq$, $\noteq$,
$>$).
Moreover, the quantification in \langnameabr is restricted to the
indices of events (so as to quantify the time points).
The idea of condition expressions is to capture a certain property of
(partial) traces. To give some intuition, before we formally define
the language, consider the ping-pong behaviour that can be specified
as follows:
%
%
\begin{center}
$
\begin{array}{r@{}c@{}l}
  \condpingponga =   \fexists i . (&i > \curr &\fand \eventquery{i}{org:resource} \noteq
                                      \eventquery{i+1}{org:resource}  
                                                \fand 
  \\
                                   &i+1 \leq \last &\fand \eventquery{i}{org:group} \eq
                                                \eventquery{i+1}{org:group}
                                      )
\end{array}
$
\end{center}

\noindent
where ``$\eventquery{i+1}{org:group}$'' is an expression for getting
the ``org:group'' attribute value of the event at the index $i+1$. The
formula $\condpingponga$ basically says that ``\textit{there exists a
  time point i that is bigger than the current time point (i.e., in
  the future), in which the resource (the person in charge) is
  different with the resource at the time point $\mathit{i+1}$ (i.e.,
  the next time point), their groups are the same, and the next time
  point is still not later than the last time point}''.
%
As for the target expression, some simple examples would be some
strings such as ``Ping-Pong'' and ``Not Ping-Pong''. Based on these,
we can create an example of \analrule 
\begin{center}
$
  \ar_1 = \ruletup{ \condpingponga \targetarrow \mbox{``Ping-Pong''}, ~ \mbox{``Not Ping-Pong''} },
$
\end{center}
where $\condpingponga$ is as above.  In this case, $\ar_1$ specifies a
task for predicting the ping-pong behaviour. In the prediction model
creation phase, we will create a classifier that classifies (partial)
traces based on whether they satisfy $\condpingponga$ or not.
During the prediction phase, such classifier can be used to predict
whether a given (partial) trace will lead into ping-pong
behaviour or not.
 
The target expression can be more complex than merely a string. For
instance, it can be an expression that involves arithmetic operations
over numeric values such as
\begin{center}
$
\begin{array}{l}
  \targetremtime = \eventquery{\last}{time:timestamp} - \eventquery{\curr}{time:timestamp},
\end{array}
$
\end{center}
which computes ``\textit{the time difference between the timestamp of
  the last event and the current event (i.e., remaining processing
  time)}''. Then we can create an \analrule
\begin{center}
$\ar_2 = \ruletup{ \curr < \last \targetarrow \targetremtime, ~0}$, 
\end{center}
%
which specifies a task for predicting the remaining time, because
$\ar_2$ will map each (partial) trace into its remaining processing
time. In this case, we will create a regression model for predicting
the remaining processing time of a given (partial) trace.
%
%
\Cref{sec:showcase} provides more examples of prediction tasks
specification using our language.




%
\medskip
\noindent
\textbf{Formalizing the Condition and Target Expressions.}
%
%
%
%
As we have seen in the examples 
above, we need to refer to a particular index of an event within a
trace. To capture this, we introduce the notion of \emph{index
  expression} $\idx$ defined as follows:
\begin{center}
$
\begin{array}{l@{ }c@{ }l}
  \idx &~::=~& i ~\mid~ \nat ~\mid~ \last ~\mid~ \curr ~\mid~ \idx_1 + \idx_2 ~\mid~ \idx_1 - \idx_2\\
\end{array}
$
\end{center}
where
\begin{inparaenum}[\itshape (i)]
\item $i$ is an \emph{index variable}.
\item $\nat$ is a positive integer (i.e., $\nat \in \mathbb{Z}^+$).
\item $\last$ and $\curr$ are special indices in which the former
  refers to the index of the last event in 
  a trace, and the latter refers to the index of the current event
  (i.e., last event of the trace prefix 
  under consideration). For instance, given a $k$-length prefix
  $\prefix{k}{\trace}$ of the trace $\trace$, $\curr$ is equal to $k$
  (or $\card{\prefix{k}{\trace}}$), and $\last$ is equal to
  $\card{\trace}$.
\item $\idx + \idx$ and $\idx - \idx$ are the usual arithmetic addition and
  subtraction operation over indices.
\end{inparaenum}


The semantics of index expression is defined over $k$-length trace
prefixes. Since an index expression can be a variable, given a
$k$-length trace prefix $\prefix{k}{\trace}$ of the trace $\trace$, we
first introduce a \emph{variable valuation} $\val$, i.e., a mapping
from index variables into $\mathbb{Z}^+$.
%
%
Then, we assign meaning to index expression by associating to
$\prefix{k}{\trace}$ and $\val$ an \emph{interpretation function}
$\inter{\cdot}{\trace}{k}{\val}$ which maps an index expression into
$\mathbb{Z}^+$. Formally, $\inter{\cdot}{\trace}{k}{\val}$ is
inductively defined as follows:
\begin{center}
$\begin{array}{lcl @{\ \ \quad\ \ } lcl @{\ \ \quad\ \ } lcl }
\inter{i}{\trace}{k}{\val} &=& \val(i) & 
\inter{\curr}{\trace}{k}{\val} &=& k& 
\inter{\idx_1 + \idx_2}{\trace}{k}{\val} &=&\inter{\idx_1}{\trace}{k}{\val} + \inter{\idx_2}{\trace}{k}{\val} \\
\inter{\nat}{\trace}{k}{\val} &=& \nat \in \mathbb{Z}^+ &
\inter{\last}{\trace}{k}{\val} &=& \card{\trace} &
\inter{\idx_1 - \idx_2}{\trace}{k}{\val} &=& \inter{\idx_1}{\trace}{k}{\val} - \inter{\idx_2}{\trace}{k}{\val}
\end{array}$
\end{center}
%

To access the value of an event attribute, we introduce \emph{event
  attribute accessor}, which is an expression of the form
\begin{center}
$
  \eventquery{\idx}{attName}
$
\end{center}
where 
$\text{\textit{attName}}$ is an attribute name 
and $\idx$ is an index expression.
%
To define the semantics of event attribute accessor, 
%
we extend the definition of our interpretation function
$\inter{\cdot}{\trace}{k}{\val}$ such that it interprets an event
attribute accessor expression into the attribute value of the
corresponding event at the given index. Formally,
$\inter{\cdot}{\trace}{k}{\val}$ is defined as follows:
\begin{center}
$
  \inter{\eventquery{\idx}{attName}}{\trace}{k}{\val}
  = \left\{ \begin{array}{l@{ \qquad }l}
\attval{attName}(e) &\mbox{{\normalsize if }}
                                        \inter{\idx}{\trace}{k}{\val}
                                        = i, ~
 1 \leq i \leq \card{\trace}\mbox{{\normalsize , and }}~ e = \trace(i) \\
\udefined & \mbox{{\normalsize otherwise}}
          \end{array}
\right.
$
\end{center}
%
        E.g., ``$\eventquery{i}{org:resource}$'' refers to the value
        of the attribute ``org:resource'' of the event at the position
        $i$.

The value of an event attribute can be either numeric (e.g.,
26, 3.86) or non-numeric (e.g., ``sendOrder''), and 
we might want to specify properties that involve arithmetic
operations over numeric values.
%
%
Thus, we introduce the notion of \emph{numeric expression} and
\emph{non-numeric expression} as expressions defined as follows:
\begin{center}
$\begin{array}{l@{ }c@{ }l}
  \nonnumexpb  &~::=~& \true  ~\mid~ \false ~\mid~ \str ~\mid~
                       \eventquery{\idx}{NonNumericAttribute}  \\
  \numexpb     &~::=~&\num ~\mid~  \idx ~\mid~
                       \eventquery{\idx}{NumericAttribute} \\
              &~\mid~&\numexpb_1 + \numexpb_2 ~\mid~ \numexpb_1 - \numexpb_2 
\end{array}$
\end{center}
where 
\begin{inparaenum}[\itshape (i)]
\item $\true$ and $\false$ are the usual boolean values,
\item $\str$ is the usual string,
\item $\num$ is real numbers,
\item $\eventquery{\idx}{\text{\textit{NonNumericAttribute}}}$ (resp.\
  $\eventquery{\idx}{\text{\textit{NumericAttribute}}}$) is
  event attribute accessor for accessing an attribute with non-numeric
  values (resp.\ numeric values),
\item $\numexpb_1 + \numexpb_2$ and $\numexpb_1 - \numexpb_2$ are the
  usual arithmetic operations over numeric expressions.
\end{inparaenum}

To give the semantics for \emph{numeric expression} and
\emph{non-numeric expression}, we extend the definition of our
interpretation function $\inter{\cdot}{\trace}{k}{\val}$ by
interpreting $\true$, $\false$, $\str$, and $\num$ as themselves
(e.g., $\inter{3}{\trace}{k}{\val} = 3$,
$\inter{\text{``sendOrder"}}{\trace}{k}{\val} = \text{``sendOrder"}$),
and by interpreting the arithmetic operations as usual, i.e., for the
addition operator we have 
\begin{center}
$\begin{array}{lcl}
  \inter{\numexpb_1 ~+~ \numexpb_2}{\trace}{k}{\val}
  &~=~& \inter{\numexpb_1}{\trace}{k}{\val} ~+~ \inter{\numexpb_2}{\trace}{k}{\val} \\
\end{array}$
\end{center}
%
The definition is similar for the subtraction operator. 
%
%
Note that the value of an event attribute might be undefined
$\udefined$. In this work, we define that the arithmetic operations
involving $\udefined$ give $\udefined$ (e.g.,
$26 + \udefined = \udefined$).

We are now ready to specify the notion of
\emph{event expression} as follows:
\begin{center}
$\begin{array}{l@{ }r@{ }c@{ }l}
  \eventexpshort  &~::=~& 
                          \numexpb_1 ~\acop~ \numexpb_2
                          &~\mid~ \nonnumexpb_1 ~\lcop~ \nonnumexpb_2\\
 & ~\mid~& \eventexpshort_1  ~\lcop~ \eventexpshort_2 &~\mid~\true ~\mid~ \false
             \end{array}$
\end{center}
where 
\begin{inparaenum}[\itshape (i)]
\item $\lcop$ stands for a logical comparison operator ($=$ or $\neq$).
\item $\acop$ stands for an arithmetic comparison operator ($<$, $>$,
  $\leq$, $\geq$, $=$ or $\neq$).
\end{inparaenum}
We interpret each logical/arithmetic comparison operator as usual
(e.g., $26 \geq 3$ is interpreted as true, ``receivedOrder'' $=$
``sendOrder'' is interpreted as false). 
%
%
%
It is easy to see how to extend the definition of our
interpretation function $\inter{\cdot}{\trace}{k}{\val}$ towards
interpreting event expressions, therefore 
we omit the details.
%


Finally, we are ready to define the language for specifying condition
expression, 
namely \langname (\langnameabr). An \langnameabr
formula is a First Order Logic (FOL) formula where the atoms are event
expressions and the quantification is ranging over event
indices. Syntactically \langnameabr is defined as follows:
%
%
\begin{center}
$\begin{array}{l@{ }c@{ }l}
  \varphi &~::=~ &  \eventexpshort ~\mid~ \neg \varphi ~\mid~ \varphi_1 \wedge
                  \varphi_2 ~\mid~ \varphi_1 \vee \varphi_2 ~\mid~ \varphi_1 \ra
                  \varphi_2 ~\mid~ \fforall i. \varphi ~\mid~ \fexists
                  i. \varphi 
\end{array}$
\end{center}
%
%
Where \eventexpshort is an event expression. The semantics of \langnameabr constructs is based on the usual FOL
semantics.
Formally, 
given a $k$-length trace prefix $\prefix{k}{\trace}$ of the trace
$\trace$, and index variables valuation $\val$, we extend the
definition of our interpretation function
$\inter{\cdot}{\trace}{k}{\val}$ 
as follows\footnote{We assume that variables are standardized apart, i.e., no two quantifiers bind the same variable (e.g., $\fforall i
  . \fexists i
  . (i > 3)$), and no variable
  occurs both free and bound (e.g., $(i > 5) \fand \fexists i
  . (i > 3)$). As usual in FOL, every FOE formula can
  be transformed into a semantically equivalent formula where the variables are
  standardized apart by applying some variable renaming
  \cite{Smul68}.}:
\begin{center}
$\begin{array}{l@{}l@{\ \ \ }c@{\ \ \ }l}
%
%
  \satisfyb{\trace}{k}{\val}{\neg\varphi}&= \true &\mbox{{\normalsize if }}&
                                                                \satisfyb{\trace}{k}{\val}{\varphi}
  = \false\\
  \satisfyb{\trace}{k}{\val}{\varphi_1 \fandcompact \varphi_2}&= \true &\mbox{{\normalsize
                                                                 if }}&
                                                                       \satisfyb{\trace}{k}{\val}{\varphi_1}
                                                                       = \true
                                                           \mbox{{\normalsize, \ and \ }}
                                                                     \satisfyb{\trace}{k}{\val}{\varphi_2}
                                                                       =
                                                                       \true\\
  \satisfyb{\trace}{k}{\val}{\exists i. \varphi}&= \true &\mbox{{\normalsize if }}&  \mbox{{\normalsize for some }} c
                                                            \in \set{1,
                                                            \ldots,
                                                            \card{\trace}}
                                                            \mbox{{\normalsize, we
                                                               have }}
                                                               \satisfyb{\trace}{k}{\val[i
                                                               \mapsto
                                                                       c]}{\varphi} = \true\\
  \satisfyb{\trace}{k}{\val}{\forall i. \varphi} &= \true&\mbox{{\normalsize if }}&  \mbox{{\normalsize for every }} c
                                                               \in \set{1,
                                                               \ldots,
                                                               \card{\trace}}
                                                               \mbox{{\normalsize, we have
                                                               that }}
                                                               \satisfyb{\trace}{k}{\val[i
                                                                       \mapsto
                                                                       c]}{\varphi}
                                                                       = \true
\end{array}$
\end{center}

\noindent
note that $\val[i \mapsto c]$ stands for a new index variable
valuation obtained from $\val$ as follows:
\begin{center}
$\val[i \mapsto c](x)
  = \left\{ \begin{array}{l@{ \qquad}l}
              c &\mbox{{\normalsize if  }} x = i\\
              \val(x) &\mbox{{\normalsize if }} x \neq i
          \end{array}
\right.$
\end{center}
Intuitively, $\val[i \mapsto c]$ substitutes each variable $i$ with
$c$, while the other variables are substituted the same way as $\val$
is defined.
The semantics of $\varphi_1 \forcompact \varphi_2$ and
$\varphi_1 \fimplcompact \varphi_2$ is as usual in FOL.
When $\varphi$ is a closed formula, its truth value does not depend on
the valuation for the index variables, and we denote the
interpretation of $\varphi$ simply by
$\inter{\varphi}{\trace}{k}{}$. We also say that \emph{$\prefix{k}{\trace}$
satisfies $\varphi$}, written $\prefix{k}{\trace} \models \varphi$, if
$\inter{\varphi}{\trace}{k}{} = \true$.


Finally, the condition expression 
in \analrules is specified as
closed \langnameabr formulas, while the target expression 
is
specified as either numeric expression or non-numeric expression,
%
except that 
target expressions are not allowed to have index variables (Thus, they
do not need variable valuation).

Essentially, \langnameabr has the following main features:
\begin{inparaenum}[\itshape (i)]
\item it allows us to specify constraints over the data;
\item it allows us to (universally/existentially) quantify different
  event time points and to compare different event attribute values at
  different event time points;
\item it supports arithmetic expressions/operations over the data.
\end{inparaenum}


\medskip
\noindent
\textbf{Checking Whether a Condition Expression is Satisfied. \xspace} 
%
%
%
%
Given a $k$-length trace prefix $\prefix{k}{\trace}$ of the trace
$\trace$, and a condition expression $\varphi$ (which is expressed as
an FOE formula), to explain how to check whether
$\prefix{k}{\trace} \models \varphi$, we first introduce some
properties of FOE formula below.  Let $\varphi$ be an FOE formula, we write
$\varphi[i \mapsto c]$ to denote a new formula obtained by
substituting each variable $i$ in $\varphi$ by $c$.

\begin{theorem}\label{thm:exist-elimination}
  Given an FOE formula $\fexists i.\varphi$, and a $k$-length trace
  prefix $\prefix{k}{\trace}$ of the trace $\trace$,
\begin{center}
$\prefix{k}{\trace} \models \fexists i.\varphi 
\mbox{ \ iff \ }
%
  \prefix{k}{\trace} \models \bigvee_{c \in \set{1, \ldots \card{\trace}}} \varphi[i \mapsto c]
$\end{center}
\end{theorem}
\begin{proof}[sketch]
  By the semantics definition, $\prefix{k}{\trace}$ satisfies
  $\fexists i.\varphi $ iff there exists an index
  $c \in \set{1, \ldots, \card{\trace}}$, such that
  $\prefix{k}{\trace}$ satisfies the formula $\psi$ that is obtained
  from $\varphi$ by substituting each variable $i$ in $\varphi$ with
  $c$. Thus, it is the same as satisfying the disjunction of formulas
  that is obtained by considering all possible substitutions of the
  variable $i$ in $\varphi$ (i.e.,
  $\bigvee_{c \in \set{1, \ldots \card{\trace}}} \varphi[i \mapsto
  c]$). This is the case because such disjunction of formulas will be
  satisfied by $\prefix{k}{\trace}$ when there is a formula in the
  disjunction that is satisfied by $\prefix{k}{\trace}$. \qed
\end{proof}

\begin{theorem}\label{thm:forall-elimination}
  Given an FOE formula $\fforall i.\varphi$, and a $k$-length trace
  prefix $\prefix{k}{\trace}$ of the trace $\trace$,
\begin{center}
$\prefix{k}{\trace} \models \fforall i.\varphi 
\mbox{ \ iff \ }
%
  \prefix{k}{\trace} \models \bigwedge_{c \in \set{1, \ldots \card{\trace}}} \varphi[i \mapsto c]
$\end{center}
\end{theorem}
\begin{proof}[sketch]
  Similar to \Cref{thm:exist-elimination}, except that we use
  conjunctions of formulas.  \qed
\end{proof}


To check whether $\prefix{k}{\trace} \models \varphi$, we perform the
following three steps:
\begin{inparaenum}[(1)]
\item Eliminate all quantifiers. This can be easily done by applying
  \Cref{thm:exist-elimination,thm:forall-elimination}. As a result,
  each variable will be instantiated with a concrete value.
\item Evaluate each event attribute accessor expression based on the
  event attributes in $\trace$. From this step, we will have a formula
  which is constituted by only concrete values composed by
  logical/comparison/arithmetic operators.
\item Last, we evaluate all logical, arithmetic and comparison
  operators.
\end{inparaenum}



\medskip
\noindent
\textbf{Formalizing the \AnalRule.} 
With this machinery in hand, now we can formalize the semantics of
\analrules as introduced above. Formally, given an \analrule
\begin{center}
$\ar = \ruletup{ 
    \cond_1 \targetarrow \target_1, ~
    \ldots, ~
    \cond_n \targetarrow \target_n, ~
    \otarget
  }.
$
\end{center}
$\ar$ is interpreted as a function that maps (partial) traces into the
values obtained from evaluating the target expressions defined below
\begin{center}
$
\ar(\prefix{k}{\trace})
  = \left\{ \begin{array}{l@{ \qquad}l}
\inter{\target_1}{\trace}{k}{} &\mbox{{\normalsize if } } \prefix{k}{\trace} \models \cond_1 \mbox{, }\\
\ \ \ \ \ \ \ \ \ \vdots& \ \ \ \ \ \ \ \  \ \ \ \ \ \ \ \ \vdots\\
\inter{\target_n}{\trace}{k}{} &\mbox{{\normalsize if } } \prefix{k}{\trace} \models \cond_n \mbox{, }\\
\inter{\otarget}{\trace}{k}{} & \mbox{{\normalsize otherwise}}
          \end{array}
\right.
$
\end{center}
where $\prefix{k}{\trace}$ is $k$-length trace prefix of the trace
$\trace$, and recall that $\inter{\target_i}{\trace}{k}{}$ is the
application of our interpretation function $\inter{\cdot}{\trace}{k}{}$
to the target expression $\target_i$ in order to evaluate the
expression and get the value. Checking whether
$\sat{\trace}{k}{\cond_i}$ can be done as explained above.
We also require that an \analrule to be \emph{coherent}, i.e., all
target expressions of an \analrule should be either only numeric or
non-numeric expressions. An \analrule in which all of its target
expressions are numeric expressions is called \emph{numeric
  \analrule}, while an \analrule in which all of its target
expressions are non-numeric expressions is called \emph{non-numeric
  \analrule}.

Given a $k$-length trace prefix $\prefix{k}{\trace}$ and an \analrule
$\ar$, we say that \emph{$\ar$ is well-defined for
  $\prefix{k}{\trace}$} if $\ar$ maps $\prefix{k}{\trace}$ into
exactly one target value, i.e., for every condition expressions
$\cond_i$ and $\cond_j$ in which $\sat{\trace}{k}{\cond_i}$ and
$\sat{\trace}{k}{\cond_j}$, we have that
$\inter{\target_i}{\trace}{k}{} = \inter{\target_j}{\trace}{k}{}$.
The notion of well-defined can be generalized to event logs. Given an
event log $\eventlog$ and an \analrule $\ar$, we say that \emph{$\ar$
  is well-defined for $\eventlog$} if for each possible $k$-length
trace prefix $\prefix{k}{\trace}$ of each trace $\trace$ in
$\eventlog$, we have that $\ar$ is well-defined for
$\prefix{k}{\trace}$. 
This condition can be easily checked for the given event log $\eventlog$
and an \analrule $\ar$.

Note that our notion of well-defined is more relaxed than requiring
that each condition must not be overlapped, and this gives flexibility
for making a specification using our language. For instance, one can
specify several characteristics of ping-pong behaviour in a more
convenient way by specifying several \condtargetrules (i.e.,
$\cond_1 \targetarrow \mbox{``Ping-Pong''}$,
$\cond_2 \targetarrow \mbox{``Ping-Pong''},\ \ldots$) instead of using
disjunctions of these several characteristics.
%
%
From now on we only consider the \analrules that are coherent and
well-defined for the event logs under consideration. 




\subsection{Building the Prediction Model }\label{subsec:pred-model}

Given an \analrule $\ar$ and an event log $\eventlog$, if $\ar$ is a
numeric \analrule, we build a regression model. Otherwise, if $\ar$ is
a non-numeric \analrule, we build a classification model.
%
%
Note that our aim is to create a prediction function that takes
(partial) traces as inputs. Thus, 
we train a classification/regression function in which the inputs are
the features obtained from the encoding of trace prefixes in the event
log $\eventlog$ (the training data).
%
%
There are several ways 
to encode (partial) traces into input features for training a machine
learning model.
For instance, \cite{LCDDM15} studies various encoding techniques such
as index-based encoding, boolean encoding, etc. In \cite{TVLD17}, the
authors use the so-called \emph{one-hot encoding} of event names, and
also add some time features 
%
(e.g., the time increase with respect to the previous event).
In general, an encoding technique can be seen as a 
%
function $\encfunc$ that takes a trace $\trace$ as the input and
produces a set $\set{x_1,\ldots, x_m}$ of features (i.e.,
$\encfunc(\trace) = \set{x_1,\ldots, x_m}$).

In our approach, users are allowed to choose the desired encoding
mechanism
%
by specifying a set $\encset$ of preferred encoding functions
(i.e., $\encset = \set{\encfunc_1, \ldots, \encfunc_n}$).
%
This allows us to do some sort of feature engineering (note that the
desired feature engineering approach, that might help increasing the
prediction performance, can also be added as one of these encoding
functions).
%
%
%
%
The set of features of a trace is then obtained by combining all
features produced by applying each of the selected encoding functions
into the corresponding trace.
In the implementation (cf.\ \Cref{sec:implementation-experiment}), we
provide some encoding functions that can be selected in order to
encode a trace. 
%
%
%
%
%
%
%

%
\begin{algorithm} \caption{: A sketch of the algorithm for building the prediction model} \label{algo:build-pred-model}
\hspace*{0mm} \textbf{Input:} 
an \analrule $\ar$, an event log $\eventlog$, a set $\encset = \set{\encfunc_1, \ldots, \encfunc_n}$ of encoding
functions \\
%
%
\hspace*{0mm} \textbf{Output:} a prediction function $\predfunc$
\begin{algorithmic}[1]

\ForEach {trace $\trace \in \eventlog$} 
\ForEach {$k \in \set{2,\ldots, \card{\trace} - 1} $} 

\State $\prefix{k}{\trace}_{\text{encoded}}$ = $\encfunc_1(\prefix{k}{\trace}) \cup \ldots \cup \encfunc_n(\prefix{k}{\trace})$

\State targetValue = $\ar(\prefix{k}{\trace})$

\State add a new training instance for $\predfunc$, where
$\predfunc(\prefix{k}{\trace}_{\text{encoded}})$ = targetValue

\EndFor
\EndFor
\State Train the prediction function $\predfunc$ (either classification or regression function)
\end{algorithmic}
\end{algorithm}
%
%

\Cref{algo:build-pred-model} illustrates how to build the
 prediction model based on the given inputs, namely:
\begin{inparaenum}[\itshape (i)]
\item an \analrule $\ar$, 
\item an event log $\eventlog$, and 
\item a set $\encset = \set{\encfunc_1, \ldots, \encfunc_n}$ of encoding functions.
\end{inparaenum}
%
%
%
%
The algorithm works as follows: for each $k$-length trace
prefix $\prefix{k}{\trace}$ of each trace $\trace$ in the event log
$\eventlog$ (where $k \in \set{2, \ldots, \card{\trace}}$), 
we do the following:
\begin{inparaenum}
\item[In line~3,] we apply each encoding function $\encfunc_i \in \encset$
  into $\prefix{k}{\trace}$,
  and combine all 
  obtained features. 
  This step gives us the encoded trace prefix.
\item[In line~4,] we compute the expected prediction result (target
  value) by applying the analytical rule $\ar$ to
  $\prefix{k}{\trace}$.
\item[In line~5,] we add a new training instance by specifying that
  the prediction function $\predfunc$ maps the encoded trace prefix
  $\prefix{k}{\trace}_{\text{encoded}}$ into the target value computed
  in the previous step.
\item[Finally,] we train the prediction function $\predfunc$ and get
  the desired prediction function.
\end{inparaenum}

\subsection{Showcase of Our Approach: Multi-Perspective Predictive
  Analysis Service}\label{sec:showcase}

An \analrule $\ar$ specifies a particular prediction task of
interest. To specify several desired prediction tasks, we only have to
specify several \analrules, i.e., $\ar_1, \ldots, \ar_2$. Given a set
$\arset$ of \analrules, i.e., $\arset = \set{\ar_1, \ldots, \ar_2}$,
our approach allows us to construct a prediction model for each
\analrule $\ar \in \arset$. This way, we can get a
\emph{multi-perspective prediction analysis service} provided by all
of the constructed prediction models where each of them focus on a
particular prediction objective.


In \Cref{subsec:spec-language} we have seen the examples of prediction
task specification for predicting the ping-pong behaviour and the
remaining processing time. In the following, we show other examples of
specifying prediction task using our
language.

\medskip
\noindent
\textbf{Predicting unexpected behaviour.\xspace} We can specify a task
for predicting unexpected behaviour by first expressing the
characteristics of the unexpected behaviour. The condition expression
$\condpingponga$ (in \Cref{subsec:spec-language}) expresses a possible
characteristic of ping-pong behaviour. Another possible
characterization of this behaviour is shown below:
\begin{center}
$
\begin{array}{r@{\ }r@{}l@{ \ }l}
  \condpingpongb =   \fexists i . (&i > \curr \fand &
                                                      \eventquery{i}{org:resource} &\noteq
                                                                                     \eventquery{i+1}{org:resource}
                                                                                     \fand\\
                                   &i+1 \leq \last \fand& 
                                                          \eventquery{i}{org:resource} &\eq
                                                                                         \eventquery{i+2}{org:resource}\fand\\
                                   &i+2 \leq \last\fand& \eventquery{i}{org:group} &\eq
                                                                                     \eventquery{i+1}{org:group} \\
                                   & \fand& \eventquery{i}{org:group} &\eq
                                                                        \eventquery{i+2}{org:group}
                                                                        )
\end{array}
$
\end{center}
essentially, $\condpingpongb$ characterizes the condition where
``\textit{an officer transfers a task into another officer of the same
  group, and then the task is transfered back into the original
  officer}''. 
%
%
In the event log, this situation is captured by the changes of the
org:resource value in the next event, but then it changes back into
the original value in the next two events, while the values of
org:group remain the same.
%
%
We can then specify an \analrule for specifying the ping-pong
behaviour prediction task as follows:
\begin{center}
$\ar_3 = \ruletup{ \condpingponga \targetarrow \mbox{``Ping-Pong''}, 
    ~ \condpingpongb \targetarrow \mbox{``Ping-Pong''}, ~ \mbox{``Not Ping-Pong''} }.$
\end{center}
Based on \Cref{algo:build-pred-model}, during the training phase,
$\ar_3$ maps each trace prefix $\prefix{k}{\trace}$ that satisfies
either $\condpingponga$ or $\condpingpongb$ into the target value
``Ping-Pong'', and those prefixes that neither satisfy
$\condpingponga$ nor $\condpingpongb$ into ``Not Ping-Pong''. 
After the training based on this rule, we get a classifier that is
trained for distinguishing between (partial) traces that will and will
not lead into ping-pong behaviour.
This example also exhibits the ability of our language to specify a
behaviour that has multiple characteristics.

\medskip
\noindent
\textbf{Predicting next event.\xspace} The task for
predicting the next event is specified as follows:
$\ar_4 = \ruletup{ \curr + 1 \leq \last \targetarrow \eventquery{\curr +
    1}{concept:name}, ~\udefined}$. 
In the training phase, $\ar_4$ maps each $k$-length trace prefix
$\prefix{k}{\trace}$ into its next event name, because
``$\eventquery{\curr + 1}{concept:name}$'' is evaluated into the name of
the event at the index $\curr~+~1$ (i.e., 
$\card{\prefix{k}{\trace}} + 1$). 
If $k = \card{\trace}$, then $\ar_4$ maps $\prefix{k}{\trace}$
into $\udefined$ (undefined).
%
%
After the training, we get a classifier that is trained to give the next event
name of the given (partial) trace.
%


\medskip
\noindent
%
%
\textbf{Predicting the next event timestamp.\xspace} This task can be
specified as follows:\footnote{Note that timestamp can be represented
  as milliseconds since epoch (hence, it is a number).}
%
\begin{center}
$\ar_5 = \ruletup{ \curr + 1 \leq \last \targetarrow \eventquery{\curr +
    1}{time:timestamp},~ \udefined}$. 
\end{center}
$\ar_5$ maps each $k$-length trace prefix $\prefix{k}{\trace}$
into the next event
timestamp. 
Hence, we train a regression model that outputs the next event
timestamp of the given (partial) trace.





\medskip
\noindent
\textbf{Predicting SLA/business constraints compliance.\xspace} Using
FOE, we can easily specify expressive SLA conditions/business
constraints, and automatically create the corresponding prediction
model using our approach. E.g., we can specify a constraint: 
\begin{center}
$\begin{array}{ll}
  \fforall\ i. (\eventquery{i}{concept:name} = \text{``OrderCreated"}
  \fimpl \fexists\ j . (j > i \fand \\
  \hspace*{10mm}  \eventquery{j}{concept:name} =
  \text{``OrderDelivered"} \fand  \eventquery{i}{orderID} =
  \eventquery{j}{orderID} \fand \\
  \hspace*{20mm} (\eventquery{j}{time:timestamp} -
  \eventquery{i}{time:timestamp}) < 10.800.000 ))
\end{array}$
\end{center}
which essentially says ``\textit{whenever there is an event where an
  order is created, eventually there will be an event where the order
  is delivered and the time difference between the two events (the
  processing time) is less than 10.800.000 milliseconds (3 hours)}''.

\section{Implementation and Experiment}
\label{sec:implementation-experiment}


As a proof of concept, by using Java and
WEKA, we have implemented a prototype\footnote{More
  information about the implementation architecture, 
  the code, the tool, and the screencast can be found at
  {{\url{http://bit.ly/predictive-analysis}}}.
} 
that is also a ProM\footnote{ProM is an extendable framework for
  process mining (\url{http://www.promtools.org}).}  plug-in.  The
prototype includes a parser for our language and a program for
automatically processing the specification as well as building the
corresponding prediction model
based on the approach explained in
\Cref{subsec:spec-language,subsec:pred-model}.
We also provide several feature encoding functions to be selected such
as one hot encoding of attributes, time since the previous event, time
since midnight, attribute values encoding, etc.
We can also choose the desired machine learning model to be built.
%
%
%
%

Our experiments aim at showing the applicability of our approach in
automatically constructing reliable prediction models based on the
given specification.
%
The experiments were conducted using the real life event log from BPI
Challenge 2013 (BPIC 13) \cite{BPI-13-data}.
%
%
%
For the experiment, 
we use the first 2/3 of the log for the training and the last 1/3 of
the log for the testing.
In BPIC 13, the ping-pong behaviour among support teams is one of the
problems to be analyzed. Ideally a customer problem should be solved
without involving too many support teams. Here we specify a
prediction task for predicting the ping-pong behaviour by first
characterizing a ping-pong behaviour among support teams as follows:
\begin{center}
$
\begin{array}{r@{ \ }c@{}l}
  \condpingpongteam =   \fexists i . (&i > \curr &\fand \eventquery{i}{org:group} \noteq
                                      \eventquery{i+1}{org:group}  
                                                \fand 
  \\
                                   &i+1 \leq \last &\fand \eventquery{i}{concept:name} \noteq
                                                \mbox{``Queued"}
                                      )
\end{array}
$
\end{center}
Roughly, $\condpingpongteam$ says that \emph{there is a change in the
support team while the problem is not being ``Queued''}. We
then specify the following \analrule:
\begin{center}
$
  \ar_{ex1} = \ruletup{ \condpingpongteam \targetarrow \mbox{``Ping-Pong''}, ~ \mbox{``Not Ping-Pong''} }
$
\end{center}
that can be fed into our tool for obtaining the prediction
model. For this case, we automatically generate Decision
Tree and Random Forest models from that specification.
%
%
We also predict the time until the next event 
by specifying the following \analrule:
\begin{center}
$\ar_{ex2} = \ruletup{ \curr + 1 \leq \last \targetarrow \eventquery{\curr+1}{time:timestamp} - \eventquery{\curr}{time:timestamp}, 0}$
\end{center}
For this case, we automatically generate Linear Regression and Random
Forest models. 

We evaluate the prediction performance of each $k$-length prefix
$\prefix{k}{\trace}$ of each trace $\trace$ in the testing set (for
$2 \leq k < \card{\trace}$).
We use accuracy and AUC (Area Under the ROC Curve) \cite{FHT01} values
as the metrics to evaluate the ping-pong prediction. For the
prediction of the time until the next event, we use MAE (Mean Absolute
Error) \cite{FHT01}, and RMSE (Root Mean Square Error) \cite{FHT01}
values as the metrics, and
we also provide the MAE and RMSE values for the mean-based prediction
(i.e., the basic approach where the prediction is based on the mean of
the target values in the training data).
The results are summarized in \Cref{tab:eventdur-ex,tab:pingping-ex}.
We highlight the evaluation for several prediction points, namely
\begin{inparaenum}[\itshape (i)]
\item early prediction (at the 1/4 of the trace length), 
\item intermediate prediction (at the 1/2 of the trace length),  and
\item late prediction (at the 3/4 of the trace length). 
\end{inparaenum}
The column ``All'' presents the aggregate evaluation for all
$k$-length prefix where $2 \leq k < \card{\trace}$.
%
%
%

\begin{table}
\caption{The evaluation of predicting ping-pong behaviour among support teams}
\centering
\begin{tabular}{ l | l | l | l | l | l | l | l | l }
  \hline
&\multicolumn{4}{l}{Accuracy}&\multicolumn{4}{|l}{AUC value}\\
\cline{2-9}
  & Early & Mid & Late & All & Early & Mid & Late & All \\
\hline
  Decision Tree & 0.82 & 0.67&0.87 &0.77 & 
                                          0.76& 0.69 & 0.63 & 0.75\\
\hline
  Random Forest & 0.83 & 0.73 & 0.91 & 0.83 & 
                                              0.89 & 0.73 & 0.78 & 0.87 \\
\hline
\end{tabular} 
 \label{tab:pingping-ex}
 \end{table}
 \begin{table}
 \caption{The evaluation of predicting the time until the next event}
 \centering
\begin{tabular}{ l | l | l | l | l | l | l | l | l }
\hline
&\multicolumn{4}{l}{MAE (in days)}&\multicolumn{4}{|l}{RMSE (in days)}\\
\cline{2-9}
  & Early & Mid & Late & All & Early & Mid & Late & All \\
\hline
Linear Regression & 0.70 & 1.42 & 2.64 & 2.07 & 
                                              1.04 & 1.87 & 2.99 & 2.77 \\
\hline
Random Forest & 0.34 & 1.07 & 1.81 & 1.51 & 
                                              1.03 & 2.33 & 2.89 & 2.61 \\
\hline
  Mean-based Prediction & 2.42 & 2.33&2.87 &2.70 & 
                                          2.44& 2.40 & 3.16 & 2.90\\
\hline
\end{tabular}
\label{tab:eventdur-ex}
\end{table}

The AUC values in \Cref{tab:pingping-ex} show that our approach is
able to automatically produce reasonable prediction models (The AUC
values $> 0.5$).
\Cref{tab:eventdur-ex} shows that all of the automatically generated
models perform better than the mean-based prediction (the baseline).
The experiment also exhibits that the performance of our approach
depends on the machine learning model that is
generated 
(e.g., in \Cref{tab:pingping-ex}, random forest performs better than
decision tree
). Since our approach does not rely on a particular machine learning
model,
it justifies that we can simply plug in different
supervised machine learning techniques in order to get
different/better performance. In the future we plan to experiment with
deep learning approach in order to get a better accuracy. As reported
by \cite{TVLD17}, the usage of LSTM neural networks could improve the
accuracy of some prediction tasks. More experiments can be seen in \Cref{sec:experiment-extended}.


%
%
%

\section{Related Work}
\label{sec:related-work}





This work is related to 
the area of predictive analysis in business process management. In the
literature, there have been several works focusing on predicting
time-related properties of running 
processes. 
For instance, the works in~\cite{
  ASS11,RW13,PSBD14,PSBD16} focus on predicting the remaining
processing time.
%
%
The works by~\cite{SWGM14,MFE12,PVFTW12} focus on predicting delays in
process execution. 
%
The authors of~\cite{TVLD17} present a deep learning approach for
predicting the timestamp of the next event and use it to predict the
remaining cycle time. 
%
%
Looking at another perspective, the works
by~\cite{MFDG14,DDFT16,VDLMD15} focus on predicting the outcomes of a
running process.
%
%
The work by~\cite{MFDG14} introduces a framework for predicting the
business constraints compliance of a running process.
%
%
In~\cite{MFDG14}, the business constraints are formulated in
propositional Linear Temporal Logic (LTL), where the atomic
propositions are all possible events during the process
executions. 
%
%
%
Another work on outcomes prediction
is presented by~\cite{PVWFT16}, which proposes an approach for
predicting aggregate process outcomes by also taking into account the
evaluation of process risk. Related to process risks,~\cite{CDLVT15} 
proposes an approach for risks prediction.
Another stream of works tackle the problem of 
predicting the future events of a running process
(cf.~\cite{TVLD17,PSBD16,ERF17,DGMPY17,BMDB16}).
%
%
%
%

A key difference 
between those works and ours is that, instead of focusing on a
specific prediction task, 
this work 
enables us to specify and focus on various prediction
tasks. To deal with these various desired prediction tasks, we also
present a mechanism that can automatically build the corresponding
prediction models based on the given specification of prediction
tasks.



This work is also related to the works on devising specification
language.
%
Unlike the propositional LTL, which is the basis of Declare language
\cite{PV06} and typically used for specifying business constraints
over sequence of events (cf. \cite{MFDG14}), our FOE language (which
is part of our rule-based specification language) allows us not only
to specify properties over sequence of events but also to specify
properties over the data (attribute values) of the events.
Concerning data-aware specification language, the work
by~\cite{BCDDM13} introduces a data-aware specification
language by combining data querying mechanisms and temporal
logic. Such language has been used in verification of data-aware
processes systems 
(cf.~\cite{AS-ICSOC-13,AS-CORR-15,AS-JELIA-14,thesis-as-16,AS-IJCAI-15,AS-RR-12a}).
%
%
The works by \cite{DMM14,MDGM13} enrich the Declare language with data
conditions based on First-Order LTL (LTL-FO). Although those languages
are data-aware, they do not support arithmetic expressions/operations
over the data which is absolutely needed, e.g., for expressing the
time difference between the timestamp of the first and the last event.
Another interesting data-aware language is S-FEEL, which is part of
the Decision Model and Notation (DMN) standard \cite{OMG15} by OMG.
Though S-FEEL supports arithmetic expressions over the data, it does
not allow us to (universally/existentially) quantify different event
time points and to compare different event attribute values at
different event time points, which is needed, e.g., in the ping-pong
behaviour specification.
%
%
%
Importantly, our language is specifically tuned for expressing
data-aware properties based on the typical structure of event logs,
and the design is highly driven by the typical prediction tasks in
business process management.
%
%


\section{Discussion and Conclusion}
\label{sec:conclusion}






We have introduced a mechanism for specifying the desired prediction
tasks by using a rule-based language, and 
for automatically creating the corresponding prediction models
based on the given specification.  A prototype of ProM plug-in that
implements our approach has been developed and several experiments
using a real life event log confirmed the applicability of our
approach.
Future work includes the extension of the tool and the language. 
One possible extension would be to incorporate aggregate functions
such as $\fsum$ and $\fconcat$. These functions enable us to specify
more tasks such as the prediction of total cost that is based on the
sum of the cost attributes in all events. The $\fconcat$ function
could allow us to specify the prediction of the next sequence of
activities by concatenating all next activities. We would also like to
extend the language with \emph{trace attribute accessor} that allows
us to specify properties involving trace attribute values.
%
There is also a possibility to exploit existing logic-based tools such
as Satisfiability Modulo Theories (SMT) solver \cite{BSST09} for
performing reasoning tasks related to the language. 
%
%
%
%
Experimenting with other supervised machine learning techniques would
be the next step as well, 
e.g., using deep learning approach in order to improve accuracy.

%
%
%
%
%
\smallskip
\noindent
\textbf{Acknowledgement.\xspace}{{ This research has been
    supported by the Euregio IPN12 ``\textit{KAOS: Knowledge-Aware
      Operational Support}'' project, which is funded by the “European
    Region Tyrol-South Tyrol-Trentino” (EGTC) under the first call for
    basic research projects. 
    The author thanks Tri Kurniawan Wijaya for various suggestions on
    this work, and Yasmin Khairina for the implementation of some
    prototype components.
    The author acknowledges the support of the Faculty of Computer
    Science of the Free University of Bozen-Bolzano, where he is
    staying for an extended research visit.
}}


\bibliographystyle{splncs03}
\bibliography{string-medium,string-medium-b,main-bib}

\appendix

\section{More Showcases of Our Approach}\label{sec:showcase-extended}

Previously, we have seen some examples of prediction task
specification using our language (cf.\
\Cref{subsec:spec-language,sec:showcase,sec:implementation-experiment}). In
the following, more examples on prediction tasks specification are
presented.

\subsection{Predicting Process Performance}

One can consider the processes that take more than certain amount of
time as ``slow process'' while the other are considered
``normal''. Given a (partial) process execution information (trace),
it might be interesting to predict whether it will end up as a slow
process or a normal process. This prediction task can be specified
as follows: 
\begin{center}
$\ar_7 = \ruletup{\cond_{71} \targetarrow \mbox{``Slow"},~ \mbox{``normal"}}$. 
\end{center}
 where 
\begin{center}
 $\cond_{71} = (\eventquery{\last}{time:timestamp} -
 \eventquery{1}{time:timestamp}) > 18.000.000$. 
\end{center}
Essentially, $\ar_7$ states that if the total running time of a trace
is greater than 18.000.000 milliseconds (5 hours), then it is
categorized as ``slow''. In the training phase, $\ar_7$ maps each
$k$-length trace prefix $\prefix{k}{\trace}$ into the corresponding
performance category (i.e., ``slow'' or ``normal'') of the
corresponding process represented by $\trace$. This way, we get a
prediction model that is trained to predict whether a certain
(partial) trace will give slow performance, or it will be normal.

Notice that one can defined more fine-grained characteristic of
process performance. For instance, we can add one more process
characteristic into $\ar_7$ by saying that those processes that spend
less than 3 hours (10.800.000 milliseconds) are considered as
``fast''. This is specified by $\ar_8$ as follows:
\begin{center}
$\ar_8 = \ruletup{\cond_{81} \targetarrow \mbox{``Slow"},~ \cond_{82} \targetarrow \mbox{``Fast"},~\mbox{``normal"}}$. 
\end{center}
 where 
\begin{center}
 $\cond_{81} = (\eventquery{\last}{time:timestamp} -
 \eventquery{1}{time:timestamp}) > 18.000.000$. \\
 $\cond_{82} = (\eventquery{\last}{time:timestamp} -
 \eventquery{1}{time:timestamp}) < 10.800.000$. 
\end{center}

\subsection{Predicting Delay}
Delay can be defined as the condition when the actual processing time
is longer that the expected processing time. Suppose we have the
information about the expected processing time, e.g., provided by an
attribute ``expectedDuration'' of the first event, we can specify an
\analrule for predicting the delay as follows:
\begin{center}
$\ar_9 = \ruletup{\cond_{91} \targetarrow \mbox{``Delay"},~\mbox{``Normal"}}$. 
\end{center}
where
\begin{center}
 $\cond_{91} = (\eventquery{\last}{time:timestamp} -
 \eventquery{1}{time:timestamp}) > \eventquery{1}{expectedDuration}$.
\end{center}
$\cond_{91}$ essentially says that the difference between the last
event timestamp and the first event timestamp (i.e., the processing
time) is greater than the expected duration (provided by the value of
the event attribute ``expectedDuration''). $\ar_9$ maps each trace
prefix $\prefix{k}{\trace}$ into either ``Delay'' or ``normal''
depending on whether the processing time of the whole trace $\trace$
is greater than the expected processing time or not. After the
training phase using this rule, we get a classifier that is trained to
distinguish between the partial traces that (probably) will and will
not lead into a delay situation.

\subsection{Predicting SLA/Business Constraints Compliance}

We have seen some prediction task specification examples for
predicting the compliance of SLA/business constraints. Another example
of an SLA would be a requirement which states that each activity must
be finished within 2 hours. This condition can be expressed as
follows:
\begin{center}
  $
\begin{array}{r@{ \ }l}
\cond_{10} = \fforall\ i. (i + 1 &\leq \last) \fimpl\\
  &(\eventquery{i+1}{time:timestamp} -
  \eventquery{i}{time:timestamp}) < 120.000
\end{array}
$
\end{center}
We can then specify an \analrule for predicting the compliance of this
SLA as follows:
\begin{center}
$\ar_{10} = \ruletup{\cond_{10} \targetarrow
  \mbox{``Comply"},~\mbox{``Not Comply"}}$. 
\end{center}
Notice that we can express the same specification in a different way, for
instance
\begin{center}
$\ar_{10}' = \ruletup{\cond_{10}' \targetarrow
  \mbox{``Not Comply"},~\mbox{``Comply"}}$. 
\end{center}
where 
\begin{center}
  $
\begin{array}{r@{ \ }l}
\cond_{10}' = \fexists\ i. (i + 1 &\leq \last) \fand\\
  &(\eventquery{i+1}{time:timestamp} -
  \eventquery{i}{time:timestamp}) > 120.000
\end{array}
$
\end{center}
Essentially $\cond_{10}'$ states that there exists a timepoint $i$, in
which $i + 1$ is still not after the last time point and the
difference between the timestamp of the event at $i+1$ and $i$ is
greater than 120.000 milliseconds (2 hours). Using either $\ar_{10}$
or $\ar_{10}'$, our algorithm for building the prediction model (cf.\
\Cref{algo:build-pred-model}) gives us a classifier that is trained to
distinguish between the partial traces that (probably will) comply and
not comply with this SLA.

\subsection{Predicting The Next Lifecycle}

The task for
predicting the lifecycle can be specified as follows:
\begin{center}
$\ar_4 = \ruletup{ \curr + 1 \leq \last \targetarrow \eventquery{\curr +
    1}{lifecycle:transition}, ~\udefined}$. 
\end{center}
In the training phase, $\ar_4$ maps each $k$-length trace prefix
$\prefix{k}{\trace}$ into its next lifecycle, because
``$\eventquery{\curr + 1}{lifecycle:transition}$'' is evaluated into
the lifecycle information of the event at the index $\curr~+~1$
(i.e., 
$\card{\prefix{k}{\trace}} + 1$).
If $k = \card{\trace}$, then $\ar_4$ maps $\prefix{k}{\trace}$
into $\udefined$ (undefined).
%
%
After the training, we get a classifier that is trained to give the
information about the most probable next lifecycle of the given
partial trace.
%


\section{Implementation}\label{sec:implementation-extended}

\begin{figure}[hbtp]
 \centering
 \includegraphics[width=0.8\textwidth]{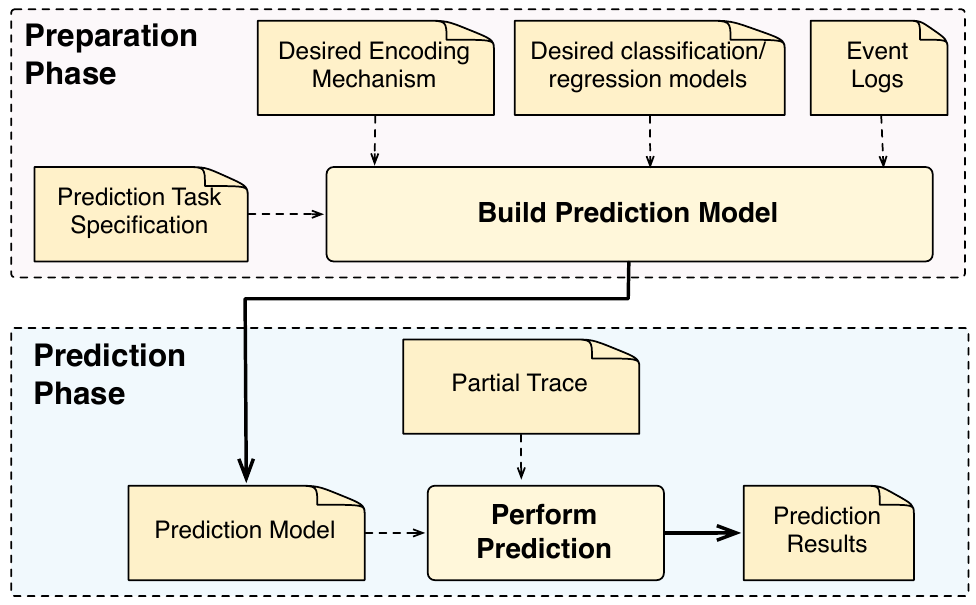}
 \caption{Specification-Driven Predictive Analysis Approach}
 \label{fig:arch}
\end{figure}

Our approach is visually described in \Cref{fig:arch}. Essentially, it
consists of two main phases, namely the preparation and the prediction
phases. In the preparation phase, we construct the prediction models
based on the given event log, as well as based on:
\begin{inparaenum}[\itshape (i)]
\item the prediction tasks specification,
\item the desired encoding mechanisms, and
\item the desired classification/regression models.
\end{inparaenum}
Once the prediction models are built, in the second phase, we can use
the generated models to perform the prediction task in order to
predict the future information of the given partial trace.

\begin{figure}[hbtp]
 \centering\hspace*{-23mm}
 \includegraphics[width=1.39\textwidth]{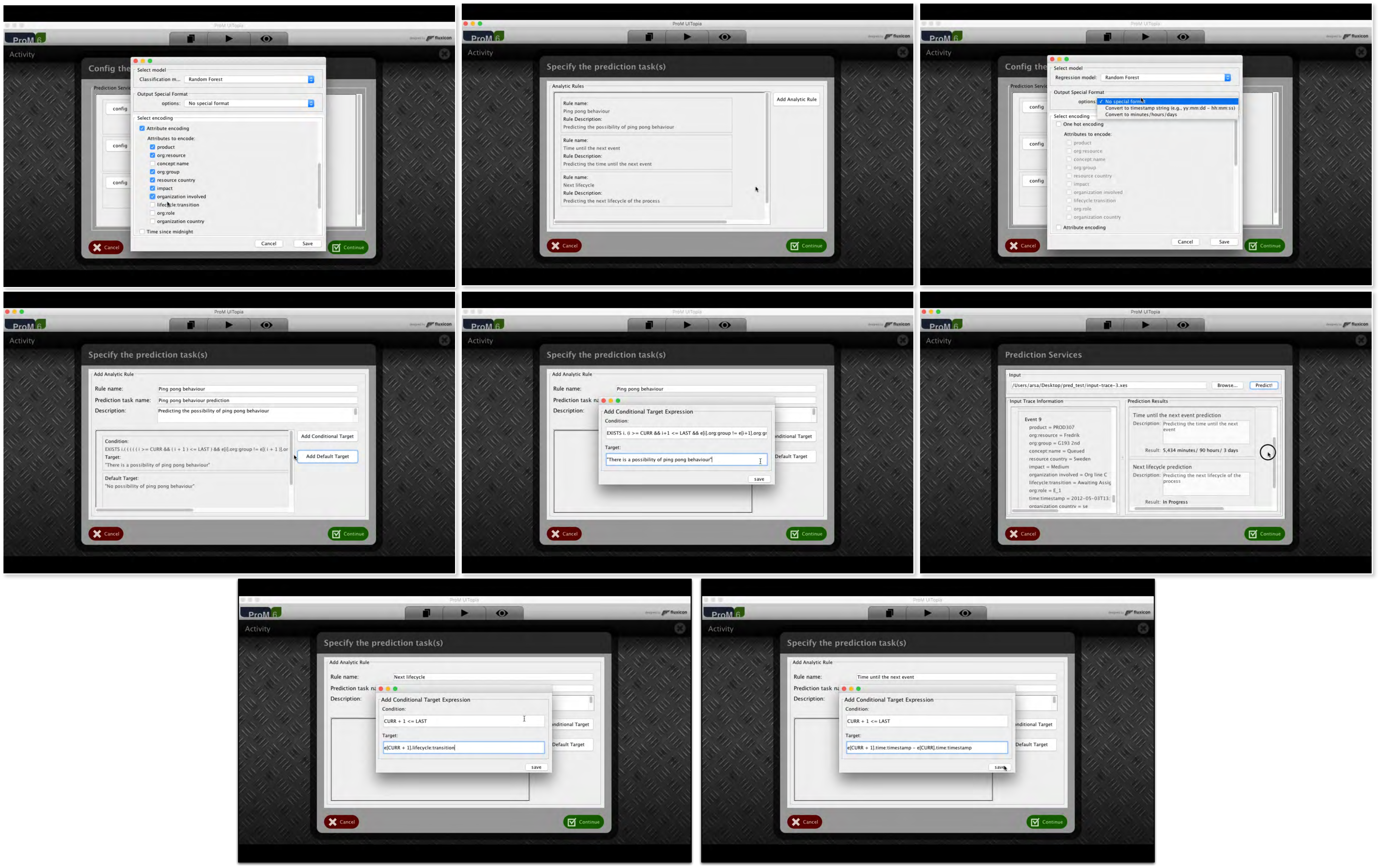}
 \caption{Some screenshots of our ProM plug-ins}
 \label{fig:prom-plugin}
\end{figure}

As a proof of concept, we have implemented two ProM plug-ins.  
One plug-in is for synthesizing the prediction models based on the
given specification, and another plug-in is for predicting the future
information of a partial trace by using the generated prediction
models. Some screenshots of our ProM plug-ins are depicted in
\Cref{fig:prom-plugin}. The screencast of our plug-ins can be found at
{{\url{http://bit.ly/predictive-analysis}}}.


\section{More Experiments}
\label{sec:experiment-extended}

We perform more experiments on the tasks that were presented in
\Cref{sec:implementation-experiment} (i.e., the prediction of
ping-pong behaviour among support teams and the time until the next
event). 
Additionally, we also conduct an experiment on another
characterization of ping-pong behaviour, namely the ping-pong
behaviour among the officers in the same group.
Apart from aiming at showing the applicability of our approach in
automatically constructing reliable prediction models based on the
given specification, we also conduct experiments on different types of
encoding techniques in order to see how it would affect the quality of
the prediction results in our approach.

One encoding technique that we use is \emph{one-hot encoding}. The
notion of one-hot encoding is as usual
(cf.~\cite{DGMPY17,TVLD17,PSBD16}), except that we are not only
considering the one-hot encoding of the activity/event name (i.e., the
attribute ``concept:name''), but we allow the possibility of adding
the one-hot encoding of another event attribute (e.g.,
``lifecycle:transition'').
In the prototype tool that we have developed, when we want to use
one-hot encoding, the users can select the corresponding event
attribute. We briefly explain the notion of one-hot encoding as
follows: Let $\textsf{\textit{attName}} \in \attnameuniv$ be an event
attribute name, $V = \set{v_1, \ldots, v_n}$ be the set of all
possible values of the attribute $\textsf{\textit{attName}}$, and
$idx(V): V \ra \set{1,\ldots, \card{V}} \subseteq \mathbb{Z}^+$ be an
ordering function over the set $V$ of all possible values of
$\textsf{\textit{attName}}$ such that $idx(v_i) = idx(v_j)$ if and
only if $v_i = v_j$ (for each $v_i, v_j \in V$). The one-hot encoding
of the attribute \textsf{\textit{attName}} that has the value $v_i$ is
a binary vector $\vec{v} \in \set{0,1}^{\card{V}}$ of length
${\card{V}}$ where all components $\vec{v}$ are set to 0, except that
the component of $\vec{v}$ at the index $idx(v_i)$ is set to 1.
%

%
We also consider the encoding 
in which we directly add the value of an event attribute
as a feature. Here, this type of encoding is called \emph{attribute
  encoding}. Similar to one-hot encoding, we can select the attribute
in which we want to apply the attribute encoding.
%
%
Note that, in some sense, the attribute encoding allows us to do an
encoding that is similar to the index-based encoding that was studied
in \cite{LCDDM15}.

\subsection{Predicting the Ping-pong Behaviour}
The experiment of predicting the ping-pong behaviour in
\Cref{sec:implementation-experiment} uses the one-hot encoding of the
attribute concept:name, the attribute encoding of ``org:resource'',
``org:group'', ``lifecycle:transition'', ``organization involved'',
``impact'', ``product'', ``resource country'', ``organization
country'', and ``org:role''.
Here we report another experiments on this problem, where we use
different types of encodings. 
The results are summarized in
\Cref{tab:pingping-12,tab:pingping-13,tab:pingping-14,tab:pingping-15,tab:pingping-17}
(For all of these experiments we add the encoding of time features,
namely the time since midnight, the time since the previous event, the
time since the first day of the week).
%
%
\begin{table}[htbp]
  \caption{The experiment of ping-pong behaviour prediction (as specified
    in \Cref{sec:implementation-experiment}) where we only use
    the one-hot encoding of the attribute concept:name}
	\centering
	\begin{tabular}{| l | l | l | l | l | l | l | l | l | }
		\hline
		&\multicolumn{4}{|c|}{Accuracy}&\multicolumn{4}{|c|}{AUC value}\\
		\cline{2-9}
		& Early & Mid & Late & All & Early & Mid & Late & All \\
		\hline
		Decision Tree & 0.73 & 0.49 & 0.93 & 0.71 & 
		0.51 & 0.30 & 0.43 & 0.67 \\
		\hline
		Random Forest & 0.72 & 0.46 & 0.76 & 0.68 & 
		0.54 & 0.36 & 0.40 & 0.68 \\
		\hline 
	\end{tabular} 
	\label{tab:pingping-12}
\end{table}
\begin{table}[htbp]
  \caption{The experiment of ping-pong behaviour prediction (as specified
    in \Cref{sec:implementation-experiment}) where we only use 
    the attribute encoding of the attribute concept:name}
	\centering
	\begin{tabular}{| l | l | l | l | l | l | l | l | l | }
		\hline
		&\multicolumn{4}{|c|}{Accuracy}&\multicolumn{4}{|c|}{AUC value}\\
		\cline{2-9}
		& Early & Mid & Late & All & Early & Mid & Late & All \\
		\hline
		Decision Tree & 0.15 & 0.81 & 0.94 & 0.77 & 
								  0.50 & 0.50 & 0.50 & 0.50 \\
		\hline
		Random Forest & 0.73 & 0.49 & 0.93 & 0.71 & 
									0.51 & 0.29 & 0.35 & 0.67 \\
		\hline
	\end{tabular} 
	\label{tab:pingping-13}
\end{table}
\begin{table}[htbp]
	\caption{The experiment of ping-pong behaviour prediction (as specified
          in \Cref{sec:implementation-experiment}) where we only use 
          the one-hot encoding of the attribute lifecycle:transition}
	\centering
	\begin{tabular}{| l | l | l | l | l | l | l | l | l | }
		\hline
		&\multicolumn{4}{|c|}{Accuracy}&\multicolumn{4}{|c|}{AUC value}\\
		\cline{2-9}
		& Early & Mid & Late & All & Early & Mid & Late & All \\
		\hline
		Decision Tree & 0.74 & 0.49 & 0.88 & 0.70 & 
								  0.51 & 0.40 & 0.54 & 0.71 \\
		\hline
		Random Forest & 0.74 & 0.48 & 0.80 & 0.69 & 
									0.56 & 0.39 & 0.50 & 0.70 \\
		\hline
	\end{tabular} 
	\label{tab:pingping-14}
\end{table}
\begin{table}[htbp]
	\caption{The experiment of ping-pong behaviour prediction (as specified
          in \Cref{sec:implementation-experiment}) where we only use 
          the attribute encoding of the attribute lifecycle:transition}
	\centering
	\begin{tabular}{| l | l | l | l | l | l | l | l | l | }
		\hline
		&\multicolumn{4}{|c|}{Accuracy}&\multicolumn{4}{|c|}{AUC value}\\
		\cline{2-9}
		& Early & Mid & Late & All & Early & Mid & Late & All \\
		\hline
		Decision Tree & 0.75 & 0.48 & 0.91 & 0.71 & 
								  0.54 & 0.35 & 0.48 & 0.69 \\
		\hline
		Random Forest & 0.74 & 0.48 & 0.80 & 0.69 & 
									0.58 & 0.40 & 0.48 & 0.70 \\
		\hline
	\end{tabular} 
	\label{tab:pingping-15}
\end{table}
\begin{table}[htbp]
	\caption{The experiment of ping-pong behaviour prediction (as specified
          in \Cref{sec:implementation-experiment}) where we use 
          the attribute encoding of the attribute org:resource,
          org:group, concept:name, lifecycle:transition, organization
          involved, impact, product, resource country, organization
          country, org:role}
	\centering
	\begin{tabular}{| l | l | l | l | l | l | l | l | l | }
		\hline
		&\multicolumn{4}{|c|}{Accuracy}&\multicolumn{4}{|c|}{AUC value}\\
		\cline{2-9}
		& Early & Mid & Late & All & Early & Mid & Late & All \\
		\hline
		Decision Tree & 0.83 & 0.67 & 0.87 & 0.78 & 
								  0.77 & 0.71 & 0.69 & 0.76 \\
		\hline
		Random Forest & 0.82 & 0.73 & 0.91 & 0.83 & 
									0.90 & 0.72 & 0.78 & 0.87 \\
		\hline
	\end{tabular} 
	\label{tab:pingping-17}
\end{table}


\vspace*{16mm}
\subsection{Predicting the Time Until the Next Event}
The experiment of predicting the time until the next event in
\Cref{sec:implementation-experiment} uses the one-hot encoding of the
attribute concept:name, the attribute encoding of ``org:resource'',
``org:group'', ``lifecycle:transition'', ``organization involved'',
``impact'', ``product'', ``resource country'', ``organization
country'', and ``org:role''.
Here we report another experiments on this problem, where we use
different types of encoding techniques. 
The results are summarized in
\Cref{tab:eventdur-22,tab:eventdur-23,tab:eventdur-24,tab:eventdur-25,tab:eventdur-27}
(For all of these experiments we add the encoding of time features,
namely the time since midnight, the time since the previous event, the
time since the first day of the week).


\begin{table}
	\caption{The experiment of time until the next event prediction (as specified
          in \Cref{sec:implementation-experiment}) where we use only
          the one-hot encoding of the attribute concept:name}
	\centering
	\begin{tabular}{| l | l | l | l | l | l | l | l | l | }
		\hline
		&\multicolumn{4}{|c|}{MAE (in days)}&\multicolumn{4}{|c|}{RMSE (in days)}\\
		\cline{2-9}
		& Early & Mid & Late & All & Early & Mid & Late & All \\
		\hline
		Linear Regression & 0.64 & 1.40 & 2.74 & 2.06 & 
										1.02 & 1.80 & 3.05 & 2.59 \\
		\hline
		Random Forest & 0.75 & 1.48 & 2.80 & 2.21 & 
									2.02 & 2.66 & 4.25 & 3.74 \\
		\hline
		Mean-based Prediction & 2.42 & 2.33 & 2.87 & 2.70 & 
												2.44 & 2.40 & 3.16 & 2.90\\
		\hline
	\end{tabular}
	\label{tab:eventdur-22}
\end{table}
\begin{table}
	\caption{The experiment of time until the next event prediction (as specified
          in \Cref{sec:implementation-experiment}) where we use only
          the attribute encoding of the attribute concept:name}
	\centering
	\begin{tabular}{| l | l | l | l | l | l | l | l | l | }
		\hline
		&\multicolumn{4}{|c|}{MAE (in days)}&\multicolumn{4}{|c|}{RMSE (in days)}\\
		\cline{2-9}
		& Early & Mid & Late & All & Early & Mid & Late & All \\
		\hline
		Linear Regression & 1.25 & 1.75 & 3.03 & 2.41 & 
										1.33 & 1.93 & 3.52 & 2.82 \\
		\hline
		Random Forest & 0.70 & 1.57 & 2.88 & 2.24 & 
									2.03 & 2.85 & 4.08 & 3.69 \\
		\hline
		Mean-based Prediction & 2.42 & 2.33 & 2.87 & 2.70 & 
												 2.44 & 2.40 & 3.16 & 2.90\\
		\hline
	\end{tabular}
	\label{tab:eventdur-23}
\end{table}
\begin{table}
	\caption{The experiment of the time until the next event prediction (as specified
          in \Cref{sec:implementation-experiment}) where we use only
          the one-hot encoding of the attribute lifecycle:transition}
	\centering
	\begin{tabular}{| l | l | l | l | l | l | l | l | l | }
		\hline
		&\multicolumn{4}{|c|}{MAE (in days)}&\multicolumn{4}{|c|}{RMSE (in days)}\\
		\cline{2-9}
		& Early & Mid & Late & All & Early & Mid & Late & All \\
		\hline
		Linear Regression & 0.61 & 1.35 & 2.30 & 1.96 & 
										0.92 & 1.77 & 2.59 & 2.56 \\
		\hline
		Random Forest & 0.70 & 1.42 & 2.61 & 2.11 & 
									1.75 & 2.48 & 3.58 & 3.35 \\
		\hline
		Mean-based Prediction & 2.42 & 2.33 & 2.87 & 2.70 & 
												 2.44 & 2.40 & 3.16 & 2.90\\
		\hline
	\end{tabular}
	\label{tab:eventdur-24}
\end{table}
\begin{table}
  \caption{The experiment of the time until the next event prediction (as specified
    in \Cref{sec:implementation-experiment}) where we use only
    the attribute encoding of the attribute lifecycle:transition}
	\centering
	\begin{tabular}{| l | l | l | l | l | l | l | l | l | }
		\hline
		&\multicolumn{4}{|c|}{MAE (in days)}&\multicolumn{4}{|c|}{RMSE (in days)}\\
		\cline{2-9}
		& Early & Mid & Late & All & Early & Mid & Late & All \\
		\hline
		Linear Regression & 0.84 & 1.36 & 2.78 & 2.22 & 
										0.94 & 1.66 & 3.34 & 2.71 \\
		\hline
		Random Forest & 0.63 & 1.36 & 2.56 & 2.10 & 
									1.70 & 2.51 & 3.68 & 3.46 \\
		\hline
		Mean-based Prediction & 2.42 & 2.33 & 2.87 & 2.70 & 
												 2.44 & 2.40 & 3.16 & 2.90\\
		\hline
	\end{tabular}
	\label{tab:eventdur-25}
\end{table}
\begin{table}
  \caption{The experiment of the time until the next event prediction (as specified
    in \Cref{sec:implementation-experiment}) where we use 
    the attribute encoding of the attribute org:resource,
    org:group, concept:name, lifecycle:transition, organization
    involved, impact, product, resource country, organization
    country, org:role}
	\centering
	\begin{tabular}{| l | l | l | l | l | l | l | l | l | }
		\hline
		&\multicolumn{4}{|c|}{MAE (in days)}&\multicolumn{4}{|c|}{RMSE (in days)}\\
		\cline{2-9}
		& Early & Mid & Late & All & Early & Mid & Late & All \\
		\hline
		Linear Regression & 0.71 & 1.44 & 2.82 & 2.15 & 
										1.02 & 1.88 & 3.22 & 2.87 \\
		\hline
		Random Forest & 0.32 & 1.11 & 2.02 & 1.61 & 
									0.84 & 2.29 & 3.04 & 2.70 \\
		\hline
		Mean-based Prediction & 2.42 & 2.33 & 2.87 & 2.70 & 
												 2.44 & 2.40 & 3.16 & 2.90\\
		\hline
	\end{tabular}
	\label{tab:eventdur-27}
\end{table}

\subsection{Predicting The Ping-pong Behaviour Among Officers in the
  Same Group}
We also experiment with another characterization of ping-pong
behaviour, namely the ping-pong behaviour among the officers in the same
group. The behaviour is specified as follows:
\begin{center}
$
\begin{array}{r@{\ }r@{}l@{ \ }l}
  \condpingpongb =   \fexists i . (&i > \curr \fand &
                                                      \eventquery{i}{org:resource} &\noteq
                                                                                     \eventquery{i+1}{org:resource}
                                                                                     \fand\\
                                   &i+1 \leq \last \fand& 
                                                          \eventquery{i}{org:resource} &\eq
                                                                                         \eventquery{i+2}{org:resource}\fand\\
                                   &i+2 \leq \last\fand& \eventquery{i}{org:group} &\eq
                                                                                     \eventquery{i+1}{org:group} \\
                                   & \fand& \eventquery{i}{org:group} &\eq
                                                                        \eventquery{i+2}{org:group}
                                                                        )
\end{array}
$
\end{center}
The results are summarized in \Cref{tab:pingping-31,tab:pingping-37}
(For all of these experiments we add the encoding of time features,
namely the time since midnight, the time since the previous event, the
time since the first day of the week).  .



\begin{table}
  \caption{Ping-pong behaviour among people in the same group, where
    we use one-hot encoding of the
    attribute concept:name, and attribute encoding of ``org:resource'',
    ``org:group'', ``lifecycle:transition'', ``organization involved'',
    ``impact'', ``product'', ``resource country'', ``organization
    country'', ``org:role''.}
\centering
\begin{tabular}{| l | l | l | l | l | l | l | l | l | }
  \hline
&\multicolumn{4}{|c|}{Accuracy}&\multicolumn{4}{|c|}{AUC value}\\
\cline{2-9}
  & Early & Mid & Late & All & Early & Mid & Late & All \\
\hline
  Decision Tree & 0.95 & 0.97 & 0.97 & 0.97 & 
                                          0.57 & 0.67 & 0.61 & 0.63 \\
\hline
  Random Forest & 0.96 & 0.99 & 0.99 & 0.99 & 
                                              0.71 & 0.79 & 0.90 & 0.81 \\
\hline
\end{tabular} 
 \label{tab:pingping-31}
 \end{table}
\begin{table}[h]
  \caption{Ping-pong behaviour among the people in the same group, where we use 
    the attribute encoding of the attribute org:resource,
    org:group, concept:name, lifecycle:transition, organization
    involved, impact, product, resource country, organization
    country, org:role}
	\centering
	\begin{tabular}{| l | l | l | l | l | l | l | l | l | }
		\hline
		&\multicolumn{4}{|c|}{Accuracy}&\multicolumn{4}{|c|}{AUC value}\\
		\cline{2-9}
		& Early & Mid & Late & All & Early & Mid & Late & All \\
		\hline
		Decision Tree & 0.96 & 0.98 & 0.98 & 0.98 & 
		0.80 & 0.77 & 0.75 & 0.78 \\
		\hline
		Random Forest & 0.96 & 0.99 & 0.99 & 0.99 & 
		0.74 & 0.87 & 0.92 & 0.84 \\
		\hline
	\end{tabular} 
	\label{tab:pingping-37}
\end{table}

\subsection{Observations}

For the prediction of the time until the next event, the results in
\Cref{tab:eventdur-27,tab:eventdur-ex} show us that encoding the value
of the attribute ``concept:name'' using one-hot encoding gives us a
better performance. However, it is not the case for the prediction of
ping-pong behaviour among officers in the same group.
\Cref{tab:pingping-31,tab:pingping-37} show that the encoding of the
``concept:name'' value using attribute encoding gives us a slightly
better performance (in particular, see the AUC values). For the
prediction of ping-pong behaviour among support teams, the results do
not show a big different whether we encode the value of the attribute
``concept:name'' using one-hot encoding or attribute encoding
(cf.~\Cref{tab:pingping-ex,tab:pingping-17}).
On the other hand, looking at
\Cref{tab:pingping-12,tab:pingping-13,tab:pingping-14,tab:pingping-15}
as well as \Cref{tab:pingping-ex,tab:pingping-17}, we see that the
more features that we use, the better performance that we
get. However, it is not always be the case, as it is exhibited by the
experiments on predicting the time until the next event.
A conclusion that we can draw from these facts is that the choice of
encodings certainly influence the quality of the prediction, and based
on those experiments, there is no encoding that is always better than
the others.
%
%
%
For instance, although one-hot encoding seems to give more
fine-grained information in the encoding, the experiments show that
encoding the information using one-hot encoding does not always give a
better result. Hence, it justifies the importance of our approach for
allowing the user to choose the desired encoding mechanism that is
more suitable for their problem.

Additionally, the experiments also show that, with a suitable choice
of encoding mechanisms, our approach is able to automatically
synthesize reliable prediction models based on the given specification
(i.e., for the classification case, the AUC values are greater than
0.5, and for the regression case, the results are better than the
mean-based prediction). Thus, it confirms the applicability of our
proposed approach in performing predictive process monitoring tasks
based on the given specification.

\end{document}